\documentclass[runningheads]{llncs}

\usepackage{caption}
\usepackage{xcolor}
\usepackage{booktabs}
\usepackage{url}
\usepackage[T1]{fontenc}
\usepackage{lmodern}
\usepackage{amsmath, amssymb}
\usepackage{graphicx}
\usepackage{array}
\usepackage{float}
\usepackage{multirow}
\usepackage{subcaption}

\begin{document}

\title{Grasynda: Graph-based Synthetic Time Series Generation}

\titlerunning{Grasynda: Graph-based Synthetic Time Series Generation}
 
\author{Luis~Amorim\inst{1,2}, Moisés~Santos\inst{2,3}, Paulo J. Azevedo\inst{1,5}, Carlos~Soares\inst{2,3,4}, Vitor~Cerqueira\inst{2,3}}

\authorrunning{Amorim et al.}

\institute{Department of Informatics, University of Minho, Braga, Portugal\\ 
\email{up202402643@edu.fe.up.pt} \and
Faculdade de Engenharia da Universidade do Porto, Porto, Portugal \and
Laboratory for Artificial Intelligence and Computer Science (LIACC), Portugal \and
Fraunhofer Portugal AICOS, Portugal\\
\and
HASLab - INESCTEC\\ 
}

\maketitle
    
    \begin{abstract}
    
    Data augmentation is a crucial tool in time series forecasting, especially for deep learning architectures that require a large training sample size to generalize effectively. However, extensive datasets are not always available in real-world scenarios. 
    Although many data augmentation methods exist, their limitations include the use of transformations that do not adequately preserve data properties.
    This paper introduces \texttt{Grasynda}, a novel graph-based approach for synthetic time series generation that: (1) converts univariate time series into a network structure using a graph representation, where each state is a node and each transition is represented as a directed edge; and (2) encodes their temporal dynamics in a transition probability matrix. 
    We performed an extensive evaluation of \texttt{Grasynda} as a data augmentation method for time series forecasting. We use three neural network variations on six benchmark datasets. The results indicate that \texttt{Grasynda} consistently outperforms other time series data augmentation methods, including ones used in state-of-the-art time series foundation models. The method and all experiments are publicly available.

    \end{abstract}
    
    \section{Introduction}
    
    In time series forecasting problems, deep learning approaches are increasingly used as effective alternatives to well-established methods such as ARIMA or exponential smoothing~\cite{bandara2020improvingaccuracyglobalforecasting,ansari2025chronos}.
    While neural networks, such as Chronos~\cite{ansari2025chronos}, NHITS~\cite{nhits} or PatchTST~\cite{cerqueira2025modelradar}, have shown state-of-the-art forecasting performance in benchmark datasets, their effective application requires a sufficiently large training sample size~\cite{bandara2020improvingaccuracyglobalforecasting,ansari2024chronos}. To address this limitation, data augmentation techniques are often used to enrich the training set by generating artificial but realistic time series. Indeed, foundation models for time series, such as Amazon's Chronos~\cite{ansari2024chronos,ansari2025chronos}, are typically trained on a mix of real and synthetic datasets to maximize their performance.

    A variety of methods have been proposed for generating synthetic time series data, ranging from simple transformations such as jittering~\cite{victor2024enhancing}, to more advanced techniques such as pattern mixing (i.e., averaging multiple time series)~\cite{forestier2017generating} or generative models~\cite{wen2020time}. Despite their popularity, these methods face important limitations. Basic augmentations such as jittering or scaling~\cite{victor2024enhancing} typically fail to either maintain temporal dependencies or to introduce sufficiently realistic variability. On the other hand, more sophisticated generative models, such as generative adversarial networks~\cite{wen2020time}, require considerable computational resources and are prone to issues such as mode collapse or overfitting, especially when training data is limited.

    \begin{figure}[b]
    \centering
    \includegraphics[width=0.35\textwidth]{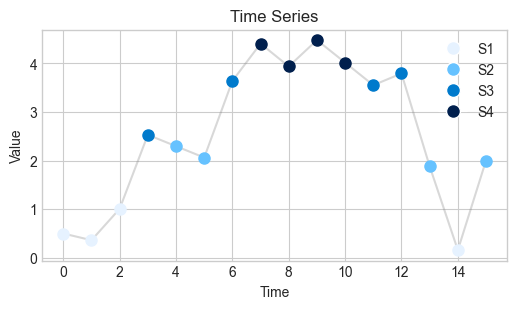}
    \includegraphics[width=0.25\textwidth]{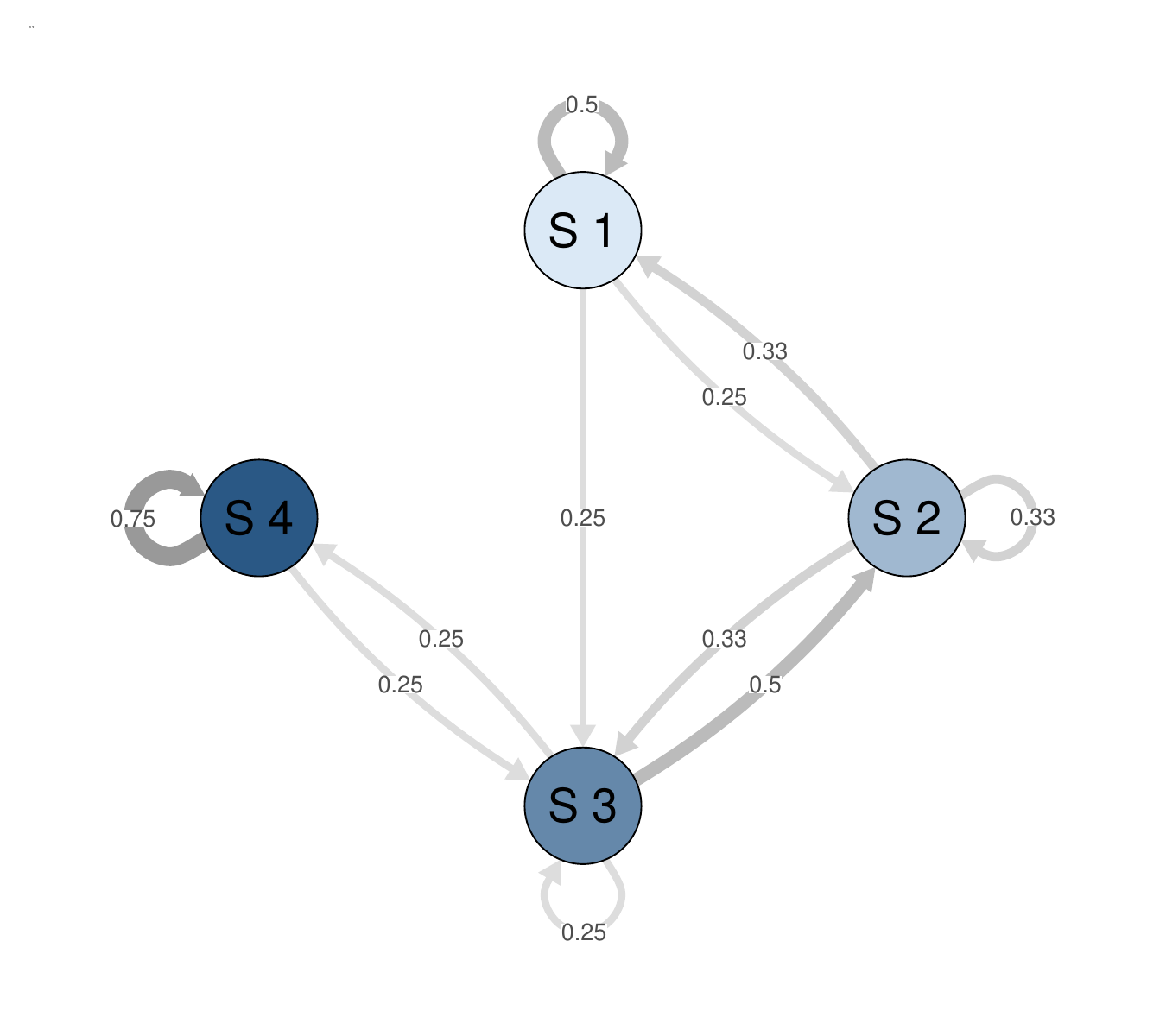}
    \includegraphics[width=0.2\textwidth]{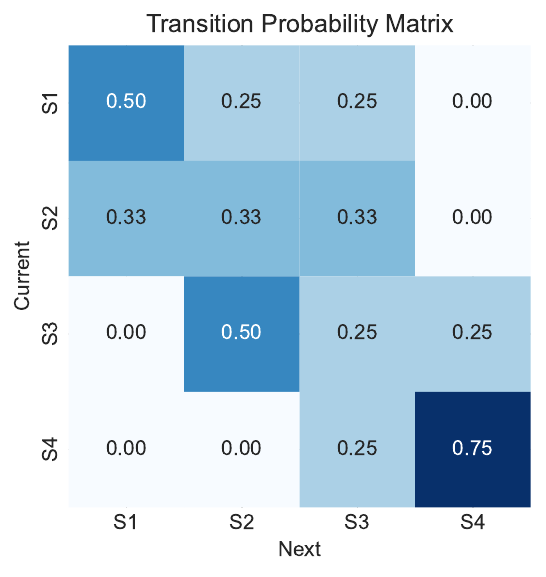}
    \caption{Example plot of a time series in the time domain (left); its network representation (middle); and the corresponding transition matrix (right).}
    \label{fig:exampleONE}
    \end{figure}
    
    To address these limitations, we introduce \texttt{Grasynda} (GRAph-based SYNthetic time series DAta generation), a novel method that generates synthetic time series by transforming them into graph representation~\cite{Silva_2021}. In this approach, the time series is discretized into a finite set of states, where each observed transition between consecutive states is represented as a directed edge, effectively capturing the temporal dynamics of the series. 
    The working hypothesis is that the network conversion framework encodes the temporal dynamics into a transition probability matrix that effectively captures the underlying data generation process. The graph representation inherently captures both local patterns (immediate state transitions) and global structures (overall network connectivity patterns). \texttt{Grasynda} leverages this matrix to generate new synthetic series by sampling transitions according to the estimated probabilities, thus maintaining the statistical properties of the original data. Figure~\ref{fig:exampleONE} illustrates this process, showing the original time series, its graph representation, and the corresponding transition matrix.
    
    We validate the usefulness of the synthetic time series created by \texttt{Grasynda} based on a forecasting downstream predictive task. We conduct experiments using six benchmark datasets and three neural network architectures. 
    The results suggest that augmenting time series datasets using \texttt{Grasynda} improves the forecasting accuracy of different neural networks. The proposed method also shows competitive forecast accuracy when compared with other data augmentation methods. The experiments and method are publicly available online\footnote{\url{https://github.com/Amorim009/Grasynda}}.

    In summary, the contributions of this paper are a novel graph-based method for synthetic time series generation, and an empirical evaluation of its effectiveness based on state-of-the-art neural network architectures applied to benchmark datasets.

    \section{Background}\label{sec:background}
    
    \subsection{Time Series Forecasting}
    
    A univariate time series is a time-ordered set of observations  \( Y = \{y_1, y_2, \dots, y_T\} \), where each value \( y_t \in \mathbb{R} \)  corresponds to the observed state at a specific time \( t \), and \( T \)  denotes the total length of the series. Forecasting refers to predicting future values of a time series, \( y_{T+1}, \dots, y_{T+h} \), where \( h \) is the forecasting horizon. Many forecasting problems often extend beyond a single univariate series, and involve collections of time series \( \mathcal{Y} = \{Y^{(1)}, Y^{(2)}, \dots, Y^{(N)}\} \) \cite{forecastingbook}, where \( N \) is the number of time series in the collection. 
    
    Most traditional forecasting approaches train separate forecasting models for each series in the collection. This makes the capture of shared patterns and dependencies that may exist across the time series impossible. Global forecasting models \cite{bandara2020improvingaccuracyglobalforecasting} overcome this limitation by using the data from all available time series to build a forecasting model. Machine learning approaches to time series forecasting typically adopt an autoregressive modeling framework \cite{bandara2020improvingaccuracyglobalforecasting}. This approach models future values as a function of the \(l\) preceding observations (lags).
    
    %\subsection{Deep Learning Approaches}
    
    %Several types of forecasting methods exist, ranging from classical methods (e.g. ARIMA \cite{forecastingbook}) to machine learning techniques. The empirical study in this work, focuses on deep learning approaches, as they have shown state-of-the-art performance in various forecasting benchmarks \cite{m4comp,makridakis2022m5}. However, the approach can be used with any (global) forecasting algorithm.
    
    Several types of deep learning architectures have been developed for forecasting tasks, ranging from feedforward neural networks and recurrent-based architectures (e.g., LSTMs, GRUs) designed to capture temporal dependencies, to more recent Transformer-based models \cite{lim2021temporal} which leverage attention mechanisms. 
    Despite the development of these diverse architectures, models based on multi-layer perceptrons (MLPs) have consistently shown competitive forecasting performance \cite{nhits}. %MLPs are neural network architectures composed of stacks of fully connected layers.
    
    One such state-of-the-art architecture based on MLPs is NHITS \cite{nhits}. NHITS is based on stacked blocks of MLP layers connected with residual links. NHITS also combines two complementary techniques, multi-rate input sampling and hierarchical interpolation, leading to state-of-the-art forecasting performance.
    
    Time series forecasting using Kolmogorov-Arnold Networks (KANs) recently emerged as a competitive alternative to MLPs~\cite{KAN}. KAN learns interpretable transformations inspired by the Kolmogorov-Arnold representation theorem. 
    Contrary to MLPs, KANs place learnable activation functions directly on the connections of the network.

    \subsection{Data Augmentation}

    Machine learning algorithms, particularly neural networks which are known to excel in data-rich environments, generally achieve better performance with larger training sets \cite{cerqueira2022case}. 
    The need for an adequate sample size is especially important when using deep learning models. 
    %These models typically contain a large number of parameters and rely on learning feature representations directly from raw data. To capture meaningful patterns and avoid overfitting, deep learning models must be exposed to a wide variety of examples. 
    Without sufficient data, they tend to memorize the training set rather than generalize to new inputs.
    Data augmentation aims to address these challenges by generating realistic variations of the original data, increasing diversity, and enabling models to generalize better to new, unseen conditions \cite{bandara2020improvingaccuracyglobalforecasting,DAadvantages}.
    The augmentation process works by generating artificial but realistic time series that mimic the characteristics of the original data. Then, the synthetic data is added to the original dataset to increase the size of the training set.

    Data generation techniques for time series forecasting have recently received increasing attention, leading to different approaches. 
    %Despite significant advances, data generation techniques for time series forecasting still face critical limitations \cite{victor2024enhancing}. 
    Transformation methods \cite{victor2024enhancing}, including time
    or magnitude warping (\texttt{T-Warp} and \texttt{M-Warp}), scaling (\texttt{Scaling}), or jittering (\texttt{Jitter}), are easily implementable due to their simplicity and computational efficiency \cite{semenoglou2023data,victor2024enhancing}. Jittering involves adding random noise to each data point, aiming to improve model robustness to such variations. \texttt{Scaling} modifies the magnitude of the series by multiplying observations by a random factor, typically drawn from a standard Normal distribution. \texttt{T-Warp} distorts the temporal axis using smooth functions such as cubic splines, which are subsequently used to resample the series \cite{roque2024rhiots}. \texttt{M-Warp} is similar but applies these smooth distortions directly to the values of the time series, leading to smoother transformations relative to \texttt{Scaling} \cite{roque2024rhiots}. These methods are adaptable to domains such as healthcare \cite{medicalDAexample}. However, they have limitations. Most importantly, they often fail to introduce sufficient variety or accurately replicate the original data generation process, losing inherent contextual integrity. 
    
    Some synthetic generation techniques focus on mixing patterns from multiple time series. These include methods such as Moving Block Bootstrap (\texttt{MBB}) combined with seasonal decomposition \cite{bergmeir2016bagging}; Dynamic Time Warping (DTW) Barycentric Averaging (\texttt{DBA}) \cite{forestier2017generating} which averages series based on DTW; and \texttt{TSMixup} \cite{ansari2025chronos} which combines randomly sampled series segments using weighted averages. The practical impact of such methods is significant. For example, Chronos \cite{ansari2025chronos} is a state of the art foundation model by Amazon that was pre-trained with synthetic time series generated using \texttt{TSMixup}. Recently, more time series models are being trained with synthetic data.
    Although generative models address some challenges, they still require high computational cost and a significant hyperparameter tuning effort  to generate data that is minimally useful \cite{victor2024enhancing}. 
    
    \subsection{Time Series Analysis using Network Science}
    
    The application of network science to time series analysis has proven to be an effective
    approach that bridges two distinct domains: temporal data analysis and graph theory \cite{Duality}. By transforming temporal data into network representations, this approach allows the use of graph theory to reveal complex patterns, nonlinear relationships, or multi-scale dynamics that conventional time and frequency-domain analyses may fail to detect. Furthermore, it offers more intuitive visualizations of dynamic behavior, which can reveal hidden structures in temporal data and improve interpretability \cite{Silva_2021}.
    
    Using network science for time series analysis involves transforming the temporal data into a graph representation. Several approaches have been proposed for this conversion. Quantile graphs \cite{Duality}, for example, discretize the time series and represent transitions between these states as edges in a graph. Another example is visibility graphs \cite{Silva_2021}, which connect time series observations that can \textit{see} each other based on a visibility criterion.
    Silva et al. \cite{Silva_2021} overview several other methods for time series analysis using network science.

    \section{The \texttt{Grasynda} Methodology}\label{sec:methodology}
    
\begin{figure}[h]
    \centering
    \includegraphics[width=0.75\textwidth]{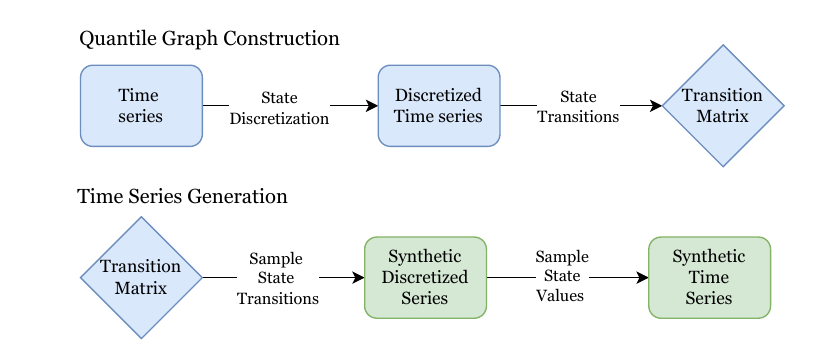}
    \caption{Workflow behind \texttt{Grasynda} for building a graph from a time series and using it to create synthetic time series.}
    \label{fig:wf}
    \end{figure}

    %This section describes the methodology behind \texttt{Grasynda}. We start by detailing the graph construction process based on quantile discretization (Section \ref{sec:qt_graph}). Section \ref{sec:genera} shows how to leverage the transition matrix to generate new time series. Then, we present a variant of \texttt{Grasynda} that combines several transition matrices for more robust time series generation (Section \ref{sec:ensemble}). Finally, we show how to use \texttt{Grasynda} to handle non-stationary time series (Section \ref{sec:nonst}).
    This proposed methodology, \texttt{Grasynda}, is illustrated in Figure~\ref{fig:wf}, consists of using a graph conversion of a time series. Section~\ref{sec:qt_graph}); Graph transitions are encoded in a matrix representation that is used as an engine for synthetic data generation. We build the probability matrix as a base for the generation process because it can be a useful tool to dynamically capture data point interactions.(Section~\ref{sec:qt_graph}).

    %\noindent The workflow behind \texttt{Grasynda}, illustrated in Figure \ref{fig:wf}, is based on two main steps: transforming time series into quantile graphs; and generating time series using this representation.

    \subsection{Time Series Graph}\label{sec:qt_graph}
    \subsubsection{Discretization.}
    
    The process begins by performing discretization on the original time series \( Y = \{y_1, y_2, \ldots, y_T\} \).
    %The series is divided into \( k \) states based on equal-frequency binning. Therefore, each bin is defined to contain an approximately equal number of observations (\( T/k \)).
    Each value \( y_t \in Y \) is mapped to a discrete label \( s_t \in \{1, 2, \ldots, k\} \) corresponding to the index of the state it belongs to. This results in a new discretized time series \( S \):
    
    \begin{equation}\label{eq:1}
        S = \{s_1, s_2, \ldots, s_T\}, \quad S_t \in \{1, 2, \ldots, k\}
    \end{equation}
    where \( s_t \) represents the corresponding label associated with the value \( y_t \). Let \( R_j \) denote the set of original values assigned to state \( s_j \):
    
    \begin{equation}
        R_j = \{ y_t \in Y \mid s_t = j \}, \quad j \in \{1, \dots, k\}
    \end{equation}
    
    \subsubsection{Graph Construction.} After discretization, we create the graph structure based on the subsequent state transitions. This structure results in a directed weighted graph \( G = (V, E, P) \), where:
    
    \begin{itemize}
        \item \( V = \{1, 2, \ldots, k\} \) is the set of nodes, each representing a state \( s_i \),
        \item \( E \subseteq V \times V \) is the set of directed edges, where an edge \(( s_i, s_j) \in E \) exists if there is at least one transition from \( s_i \) to \( s_j \) in the discretized time series \( S \),
        \item \( P = [p_{ij}] \) is the normalized edge weight matrix (transition matrix), where each weight \( p_{ij} \) corresponds to the empirical conditional probability of transitioning from state \( s_i \) to state \( s_j \):
    \end{itemize}
    The elements of \( P \) satisfy:
    
    \begin{equation}
        p_{ij} = \frac{\text{count of transitions from } s_i \text{ to } s_j}{\text{total transitions from } s_i}
    \end{equation}
    
    %\noindent Since these are conditional probabilities, each row of \( P \) sums to 1:
    
    %\begin{equation}
       % \sum_{j=1}^{k} p_{ij} = 1, \quad %\forall i \in \{1, \dots, k\}.
    %\end{equation}
    %
    
    \noindent Since these are conditional probabilities, each row of \( P \) sums to 1: \( \sum_{j=1}^{k} p_{ij} = 1 \) for all \( i \in \{1, \dots, k\} \).
    Generally $p_{ij} \neq p_{ji}$, indicating the directed nature of the graph and the non-symmetric property of the transition matrix.

    \subsection{Data generation}\label{sec:genera}
    
    %\subsubsection{Discretized Series Generation.}
    
    By sampling from the transition matrix $P$, we can generate synthetic sequences \( \hat{S} = \{\hat{s}_1, \hat{s}_2, \dots, \hat{s}_T\} \). We start by selecting an initial state \( \hat{s}_1 \), matching the first state from the original time series. The subsequent states are sampled iteratively based on the transition probabilities:
    
    \begin{equation}
        \hat{s}_{t+1} \sim P(\hat{s}_t)
    \end{equation}
    
    This means that \( \hat{s}_{t+1} \) is chosen according to the probability distribution given by the row of \( P \) corresponding to \( \hat{s}_t \). In effect, the synthetic discretized synthetic sequence follows the same probabilistic transitions as the original discrete series. We can sample any number of states from the matrix to generate synthetic sequences of arbitrary length. 
    
    %Notwithstanding, in the experiments presented in the next section, we sample a number of quantiles equal to the size of the original time series.
    
    %\subsubsection{Time Series Generation.}
    
    %Initially, random numbers within the range of each quantile bin were generated to map the artificial quantile series to time series values. However, this approach presented significant issues, as the introduction of random values introduced considerable noise, leading to a distribution that deviated from the original series. 
    %To address this problem, the method was refined 
    After obtaining a synthetic discrete time series $\hat{S}$, we convert it back to the original time series space. This is achieved by sampling from the value distribution of each corresponding state . Specifically, for each state \( \hat{s}_i \) in \( \hat{S} \), a synthetic value \( \hat{y}_i \) is sampled from its distribution. $\hat{y}_t \sim \text{Uniform}(R_{\hat{s}_i})$
    
    This results in a synthetic time series \( \hat{Y} = \{ \hat{y}_1, \hat{y}_2, \dots, \hat{y}_T \} \). This sampling approach is expected to preserve the original temporal dynamics of the time series leveraging the probabilistic nature of the graph matrix transitions.
    
    \section{Quantile-based Grasynda}\label{sec:study}
    
    In this section, we describe an instantiation of Grasynda using quantile graphs as the method for converting time series into network representations. 
    
    \subsection{Time series quantile graph}
    
    For the quantile-based version of our method, each value in the time series is assigned to a quantile using equal-frequency binning. This produces a discrete sequence with the same length as the original series, where each element corresponds to a quantile label. The sequence is defined by a directed graph in which each node represents a quantile level, and each edge captures the transitions observed between consecutive labels.
    
    We count the transitions from one quantile to the others and normalise these counts to obtain the transition probability matrix of the quantile graph. The final synthetic time series is generated by following the general Grasynda methodology described in Section \ref{sec:methodology}.

    \subsection{Data preprocessing: handling non-stationarity}\label{sec:nonst}
    
    The adopted tool to convert time series into network structures, quantile graph conversion, focuses on  transitions between consecutive observations. While this tool is adequate to capture local patterns through edge connections, it has limitations in preserving long-term temporal dependencies such as trend and seasonality that characterize non-stationary time series.
    
    To address these limitations, we adopt a decomposition-based approach following Bergmeir et al. \cite{bergmeir2016bagging}. Our solution leverages Seasonal-Trend decomposition using LOESS (STL) to decompose each time series into three distinct components:
    \begin{equation}
        y_t = \text{Trend}_t + \text{Seasonal}_t + \text{Remainder}_t
    \end{equation}
    
    We then apply the quantile graph version of \texttt{Grasynda} to the remainder component, which typically exhibits stationarity. After obtaining a synthetic remainder component $Remainder_{\texttt{Grasynda}}$ is generated, we recombine it with the original trend and seasonal components to produce the final synthetic time series:
    \begin{equation}
        \hat{y}_t = \text{Trend}_t + \text{Seasonal}_t + \text{Remainder}_{\texttt{Grasynda}}
    \end{equation}

\begin{figure}[t]
    \centering
    \includegraphics[width=0.33\textwidth]{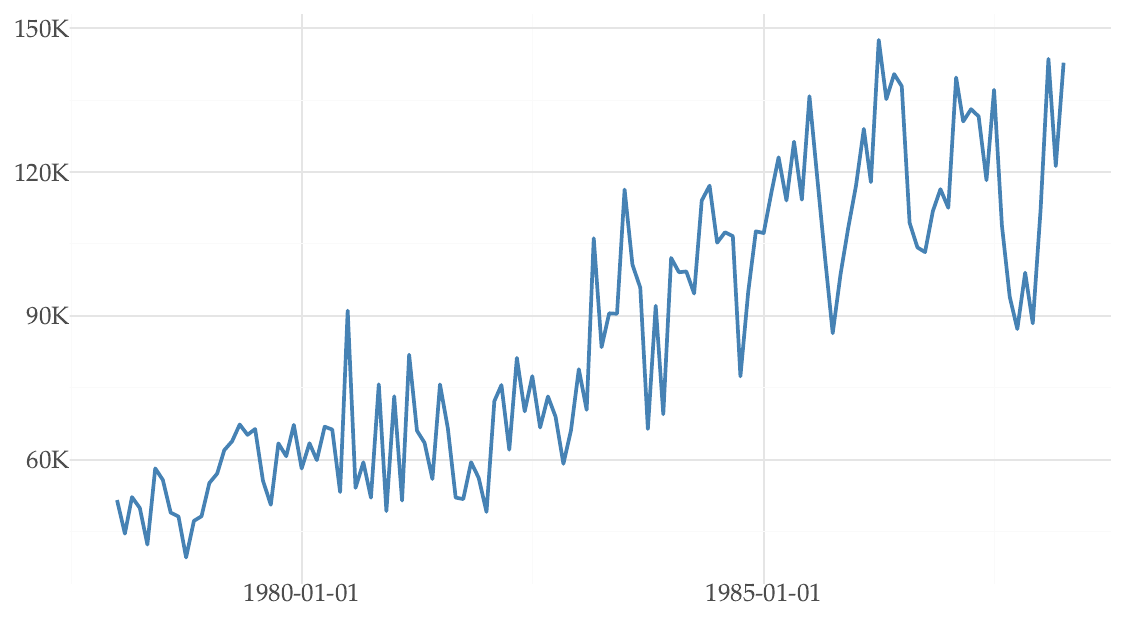}
    \includegraphics[width=0.27\textwidth]{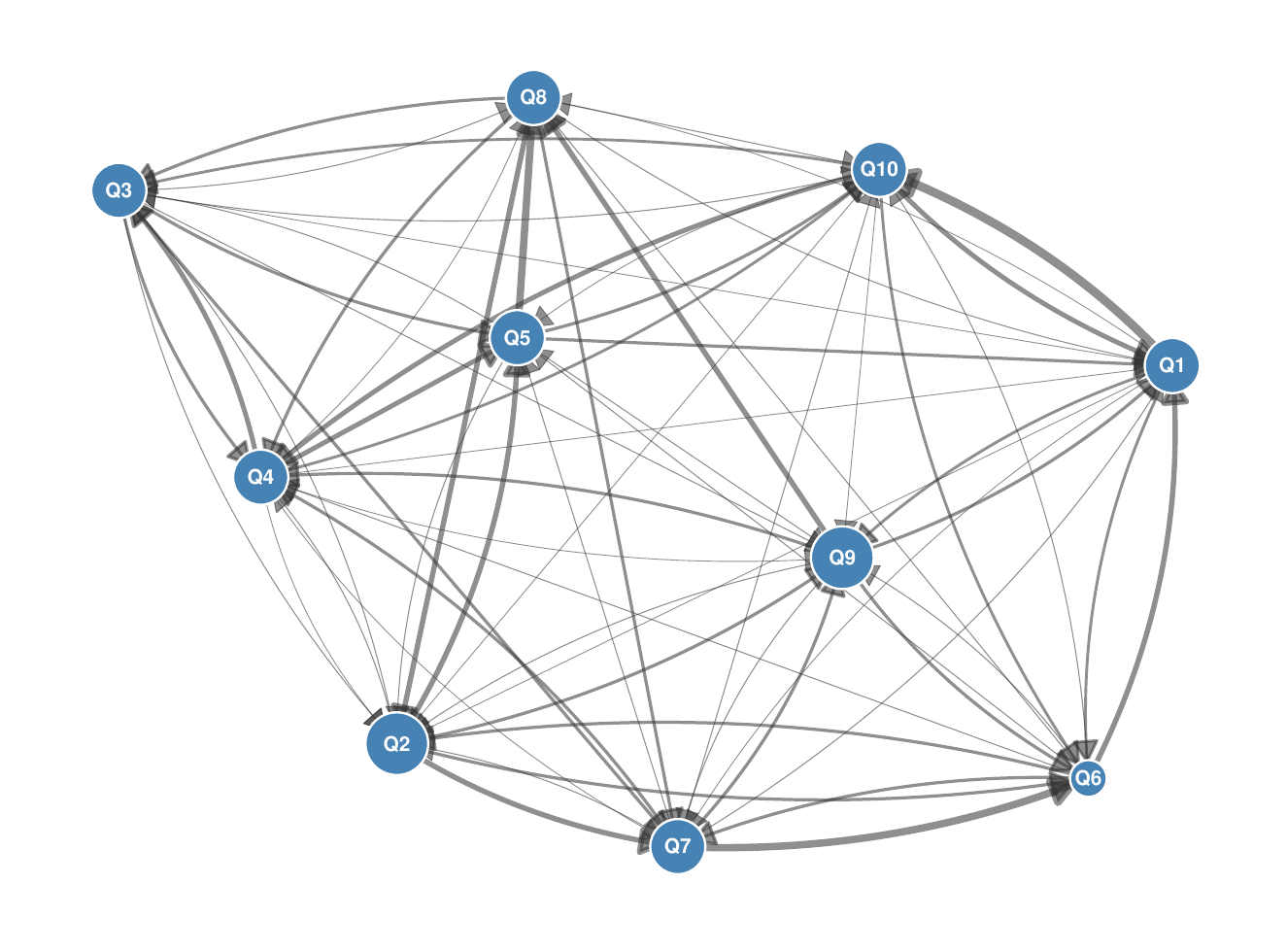}
    \includegraphics[width=0.33\textwidth]{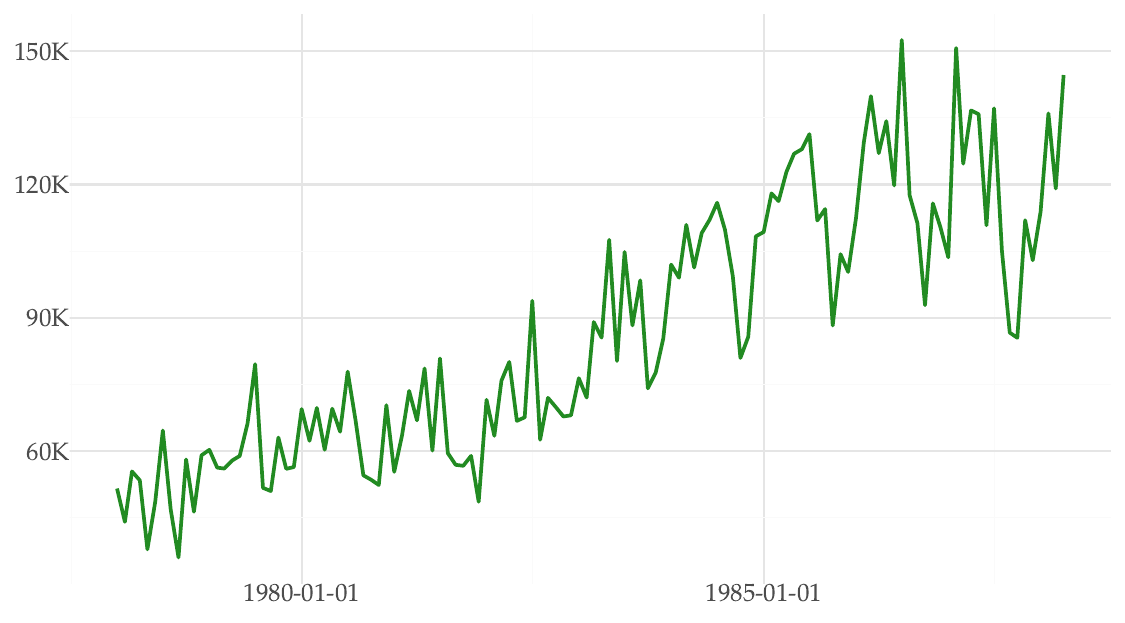}
    
    \caption{Example plot of a time series in the time domain (left); its network representation using a quantile graph (middle); and a synthetic time series generated from the graph using \texttt{Grasynda} (right).}
    \label{fig:example}
    \end{figure}

    Figure \ref{fig:example} shows an example application of \texttt{Grasynda} using the time series with identifier \textit{ID90} from the M3 Monthly dataset (c.f. Section~\ref{sec:datasets_}). The figure shows the time series in the standard time domain (a), the corresponding network representation using a quantile graph (b), and a synthetic time series generated from the graph using \texttt{Grasynda} (c). We note that this preprocessing operation may be useful for other instantiations of the methodology as well.

    \section{Experiments}\label{sec:experiments}
    
    To assess the effectiveness of the synthetic time series generated by \texttt{Grasynda}, we performed a series of experiments focused on a downstream forecasting task. Specifically, we evaluated \texttt{Grasynda} in comparison to several state-of-the-art synthetic time series generation methods, using six benchmark datasets and three neural network architectures. 
    %We address the following research questions:
    %\begin{itemize}
    %    \item \textbf{RQ1}: How does \texttt{Grasynda} compare to existing synthetic data generation techniques in terms of downstream forecasting accuracy across different neural network architectures?
    %    \item \textbf{RQ2}: How effective is the proposed graph-based representation for generating synthetic time series and augmenting training datasets to improve forecasting performance?
    %\end{itemize}
    We address the following research question: 
    \emph{How effective is \texttt{Grasynda} as a data augmentation for time series forecasting?}
    
    \subsection{Experimental Design}\label{sec:edesign}
    
    \subsubsection{Forecasting problems.}\label{sec:datasets_}
    
    We use six publicly available benchmark datasets from three forecasting competitions: M1 \cite{makridakis1982accuracy}, M3 \cite{makridakis2000m3}, and Tourism \cite{athanasopoulos2011tourism}. These datasets span various application domains, such as industry, demography, and economics. Table~\ref{tab:data} provides a detailed summary of their characteristics. In total, the datasets comprise 3,797 time series and 409,602 individual observations. For forecasting tasks, we set the horizon ($h$) to 12 for monthly time series and 8 for quarterly time series. The input window (number of lagged observations) is fixed at 24 for monthly and 8 for quarterly data, corresponding to two full seasonal cycles for each frequency.
    
    \begin{table}[h]
    \centering
    \caption{Summary of the datasets: average value, number of time series, number of observations, seasonal period, and forecasting horizon (h).}
    \label{tab:data}
    \begin{tabular}{lrrrrr}
    \toprule
    Dataset & Avg. Value & \# Series & \# Obs. & Period & h \\
    \midrule
    M1-M       & 72.7  & 617  & 44,892  & 12 & 12 \\
    M1-Q     & 40.9  & 203  & 8,320   & 4  & 8  \\
    M3-M       & 117.3 & 1428 & 167,562 & 12 & 12 \\
    M3-Q      & 48.9  & 756  & 37,004  & 4  & 8  \\
    T-M  & 298.5 & 366  & 109,280 & 12 & 12 \\
    T-Q & 99.6  & 427  & 42,544  & 4  & 8  \\
    \midrule
    Total             & -     & 3797 & 409,602 & -  & -  \\
    \bottomrule
    \end{tabular}
    \end{table}

    \subsubsection{Performance estimation.}
    
    To assess forecasting performance, we hold out the final $h$ observations of each time series as a test set, while training models on all preceding data points. Forecast accuracy is evaluated using the Mean Absolute Scaled Error (MASE), a widely adopted, scale-independent metric. MASE is calculated as:
    \(
    \text{MASE} = \frac{\text{MAE}}{\text{MAE}_{\text{naive}}}
    \),
    where \( \text{MAE} = \frac{1}{T} \sum_{t=1}^{T} |y_t - \hat{y}_t| \) denotes the mean absolute error of the forecast, and \( \text{MAE}_{\text{naive}} = \frac{1}{T - 1} \sum_{t=2}^{T} |y_t - y_{t - 1}| \) corresponds to the mean absolute error of a one-step naive forecast.
    
    Given a training dataset \( \mathcal{D} = \{ Y^{(1)}, Y^{(2)}, \ldots, Y^{(M)} \} \), where each time series is \( Y^{(i)} = \{ y^{(i)}_1, y^{(i)}_2, \ldots, y^{(i)}_T \} \), we generate a synthetic series \( \hat{Y}^{(i)} = \{ \hat{y}^{(i)}_1, \hat{y}^{(i)}_2, \ldots, \hat{y}^{(i)}_T \} \) for every \( Y^{(i)} \). The augmented training set, therefore, consists of both the original and synthetic time series:
    
    \begin{equation}
    \mathcal{D}_{\text{aug}} = \{ Y^{(1)}, \ldots, Y^{(M)}, \hat{Y}^{(1)}, \ldots, \hat{Y}^{(M)} \}
    \end{equation}

    \subsection{Methods}
    
    %\subsubsection{Neural Networks and Baseline.}
    
    We evaluate three state-of-the-art neural network forecasting models: NHITS~\cite{nhits}, KAN~\cite{KAN}, and MLP. These models were chosen for their competitive forecasting accuracy (see Section~\ref{sec:background}) and computational efficiency relative to other deep learning architectures, such as transformer-based models~\cite{forecastingModels}. We include the seasonal naive method as a baseline, providing a reference point for the experimental results. Each neural network is implemented using the neuralforecast Python library and optimized via 10 iterations of random search for hyperparameter selection.

    %\subsubsection{Synthetic time series generators.}
    
    We benchmark the proposed method against several widely used data augmentation techniques: \texttt{Scaling}~\cite{semenoglou2023data}; \texttt{TSMixup}, use by Amazon's time series foundational model Chronos~\cite{ansari2024chronos,ansari2025chronos}; \texttt{DBA}~\cite{forestier2017generating}; jittering (\texttt{Jitter})~\cite{semenoglou2023data}; magnitude warping (\texttt{M-Warp})~\cite{MagnitudeWarping}; time warping (\texttt{T-Warp})~\cite{TimeWarping}; and seasonal moving block bootstrap (\texttt{MBB})~\cite{bergmeir2016bagging}. These methods are further described in Section~\ref{sec:background}. As a baseline, we also consider results using the original, non-augmented training data (\texttt{Original}). All augmentation methods are configured according to settings established in previous literature. For our method, \texttt{Grasynda}, we use 25 quantiles.

    \subsection{Results}\label{sec:results}
    
    Table~\ref{tab:scores_by_ds} summarizes the experimental results, reporting the MASE scores for each synthetic time series generation method across different model architectures and datasets. The table also includes the average performance and average rank for each method (with lower ranks indicating better results). Asterisks ('*') indicate methods that achieve significantly better performance than \texttt{Original}, according to the Wilcoxon signed-rank test. 
    %Across all experiments, the neural network models consistently outperform the seasonal naive baseline, confirming the validity of the forecasting performance of the trained models.
    The table confirms the validity of the experimental setup, as all the neural network models consistently outperform the seasonal naive baseline.

    \begin{table}[!t]
    \caption{MASE of each approach across different architectures and datasets. Best and second-best values are \textbf{bolded} and \underline{underlined}, respectively. Asterisks (*) denote significant improvements over \texttt{Original}.}
    \label{tab:scores_by_ds}
    \centering
    \resizebox{.9\textwidth}{!}{%
    \begin{tabular}{llllllllllllll}
    \toprule
     &  & \rotatebox{60}{\texttt{Original}} & \rotatebox{60}{\texttt{Grasynda}} & \rotatebox{60}{\texttt{DBA}} & \rotatebox{60}{\texttt{Jitter}} & \rotatebox{60}{\texttt{M-Warp}} & \rotatebox{60}{\texttt{MBB}} & \rotatebox{60}{\texttt{Scaling}} & \rotatebox{60}{\texttt{T-Warp}} & \rotatebox{60}{\texttt{TSMixup}} & \rotatebox{60}{\texttt{SNaive}} \\
    \midrule
    \multirow{6}{*}{\rotatebox[origin=c]{90}{NHITS}}
     & M1-M & 0.977 & \textbf{0.945}* & \underline{0.948*} & 0.970 & 0.981 & 0.952 & \underline{0.948*} & 0.977 & 0.978 & 1.221 \\
     & M1-Q & 1.037 & 1.034 & 1.094 & 1.036 & \textbf{0.974}* & \underline{1.029} & 1.039 & 1.066 & 1.049 & 1.647 \\
     & M3-M & 0.801 & \underline{0.759}* & 0.779 & 0.786 & 0.773 & 0.764* & 0.770 & \textbf{0.757}* & 0.770 & 1.091 \\
     & M3-Q & 1.199 & 1.220 & \underline{1.136}* & 1.185 & 1.181 & 1.165* & \textbf{1.131}* & 1.193 & 1.193 & 1.417 \\
     & T-M & 1.208 & \underline{1.190}* & 1.193 & 1.200* & 1.208 & \textbf{1.184}* & 1.206 & 1.230 & \underline{1.190*} & 1.345 \\
     & T-Q & 1.635 & \textbf{1.555}* & \underline{1.565*} & 1.608* & 1.628 & 1.663 & 1.603* & 1.616 & 1.622 & 1.702 \\
    \cmidrule(lr){2-12}
     & Avg. & 1.143 & \underline{1.117} & 1.119 & 1.131 & 1.124 & 1.126 & \textbf{1.112} & 1.129 & 1.134 & 1.404 \\
     & Avg. Rank & 7.3 & \textbf{3.1} & 4.4 & 5.2 & 5.8 & \underline{3.7} & 3.8 & 6.0 & 5.8 & 10.0 \\
    \midrule
    \multirow{6}{*}{\rotatebox[origin=c]{90}{MLP}}
     & M1-M & \underline{0.930} & 0.948 & 0.985 & 0.940 & 0.968 & 0.936 & \textbf{0.926} & 0.977 & 0.954 & 1.221 \\
     & M1-Q & 1.070 & 1.075 & 1.064 & \underline{1.034}* & \textbf{0.996}* & 1.036* & 1.040 & 1.076 & 1.062 & 1.647 \\
     & M3-M & 0.774 & 0.768 & 0.777 & \underline{0.760}* & 0.780 & 0.765* & 0.767 & \textbf{0.757} & 0.761 & 1.091 \\
     & M3-Q & 1.143 & 1.181 & 1.152 & 1.148 & 1.159 & 1.154 & \textbf{1.112} & \underline{1.132} & 1.209 & 1.417 \\
     & T-M & 1.198 & \underline{1.195} & 2.545 & 1.213 & 1.216 & 1.196* & 1.204 & 1.257 & \textbf{1.183}* & 1.345 \\
     & T-Q & 1.568 & \textbf{1.517}* & 1.605 & 1.654 & 1.615 & 1.637 & 1.645 & 1.720 & 1.598 & 1.702 \\
    \cmidrule(lr){2-12}
     & Avg. & \textbf{1.114} & \textbf{1.114} & 1.355 & 1.125 & 1.122 & \underline{1.121} & 1.116 & 1.153 & 1.128 & 1.404 \\
     & Avg. Rank & \underline{4.2} & 5.0 & 7.0 & 4.3 & 6.0 & 4.2 & \textbf{3.8} & 6.3 & 4.5 & 9.7 \\
    \midrule
    \multirow{6}{*}{\rotatebox[origin=c]{90}{KAN}}
     & M1-M & 0.961 & 0.941 & 0.957 & 0.962 & 0.979 & \underline{0.939} & \textbf{0.936} & 1.008 & 0.955 & 1.221 \\
     & M1-Q & \underline{1.013} & 1.022 & 1.043 & 1.013 & \textbf{0.966}* & 1.016 & 1.030 & 1.071 & 1.044 & 1.647 \\
     & M3-M & 0.784 & 0.779 & 0.777 & 0.783 & 0.796 & \textbf{0.773}* & 0.797 & 0.798 & \underline{0.775*} & 1.091 \\
     & M3-Q & 1.223 & 1.207* & \textbf{1.146}* & 1.229 & 1.213 & 1.221 & \underline{1.165*} & 1.206 & 1.271 & 1.417 \\
     & T-M & 1.227 & \textbf{1.190}* & 1.231 & 1.228 & 1.229 & 1.217 & 1.222 & 1.273 & \underline{1.213}* & 1.345 \\
     & T-Q & 1.571 & \textbf{1.548} & \underline{1.549} & 1.631 & 1.591 & 1.600 & 1.623 & 1.670 & 1.642 & 1.702 \\
    \cmidrule(lr){2-12}
     & Avg. & 1.130 & \textbf{1.114} & \underline{1.117}& 1.141 & 1.129 & 1.128 & 1.129 & 1.171 & 1.150 & 1.404 \\
     & Avg. Rank & 4.9 & \textbf{3.0} & 4.3 & 5.9 & 5.3 & \underline{3.5} & 4.5 & 8.0 & 5.5 & 10.0 \\
    \midrule
    \multicolumn{2}{l}{Effectiveness} & -- & \textbf{0.72} & 0.56 & 0.50 & 0.39 & \underline{0.67} & \underline{0.67} & 0.33 & 0.56 & 0.0 \\
    \bottomrule
    \end{tabular}%
    }
    \end{table}
    
    Overall, \texttt{Grasynda} shows consistently competitive forecasting performance compared to other synthetic time series generation approaches across all model architectures. It achieves the lowest (best) average MASE in two out of the three evaluated forecasting models, and ranks second in the remaining model. The average rank scores, calculated for each forecasting model, further indicate that \texttt{Grasynda} outperforms other data augmentation methods.
    
    We also assessed the effectiveness of each approach in augmenting the training datasets. For each of the 18 evaluated experiments (6 datasets times 3 forecasting models), we counted the number of times each augmentation method outperformed the no-augmentation baseline (\texttt{Original}). \texttt{Grasynda} achieved better results than \texttt{Original} in 13 out of 18 cases (72\%), representing the highest effectiveness among all augmentation methods. Additionally, in 7 of these 18 tests, \texttt{Grasynda} delivered a statistically significant improvement over \texttt{Original}. By contrast, most other augmentation approaches rarely showed significant improvements over the baseline, with the exception of \texttt{MBB}, which also achieved significant gains 7 times.
    
    An important remark is the comparison between \texttt{Grasynda} and \texttt{TSMixup} \cite{ansari2025chronos}. Our model obtained a better average MASE and rank for all three architectures, suggesting that it outperforms \texttt{TSMixup}. This is particularly interesting because \texttt{TSMixup} is used to create data for training a state of the art time series foundation model, Chronos \cite{ansari2024chronos}.

    \section{Discussion}
    
    The experimental results highlight the effectiveness of \texttt{Grasynda} as a data augmentation method for a time series forecasting downstream task. Across different benchmark datasets and forecasting architectures, \texttt{Grasynda} improved forecasting accuracy over the no-augmentation baseline (\texttt{Original}) in 72\% of cases, and achieved the best or second-best average ranks among all compared methods. Its gains consistently surpassed those of state-of-the-art alternatives, including \texttt{TSMixup}, which has recently been leveraged to train state-of-the-art time series foundation models such as Chronos~\cite{ansari2024chronos,ansari2025chronos}. These findings underscore the potential of modeling time series structure using graph representations for generating realistic and useful synthetic data.
    
    Despite these promising results, some limitations and opportunities for future research remain. First, the current version of \texttt{Grasynda} is specifically tailored to univariate time series and does not explicitly address multivariate data or long-range dependencies. Extending the graph construction and generation process to handle these cases is an open opportunity. Second, our approach relies on a straightforward quantile-based discretization scheme; more sophisticated discretization strategies (e.g. visibility graphs~\cite{Silva_2021}) may further enhance its flexibility, or generalizability to diverse domains. 
    Another consideration is that the quantile-based discretization scheme used in \texttt{Grasynda} does not directly model non-stationarity sources of variation, such as trend or seasonality. As a result, adequate preprocessing, such as STL decomposition, may be necessary to ensure the generated synthetic series are realistic. In effect, another promising direction is to integrate these aspects more directly into the graph-based generation framework.
    
    Overall, the evidence shows the potential of employing graph-based approaches for synthetic time series generation, pointing toward interesting future research directions at the intersection of network science and time series modeling.

    %\section{Conclusions}
    
    %This paper introduced \texttt{Grasynda}, a novel synthetic time series generation method. By converting time series into graph representations and using transition probability matrices, \texttt{Grasynda} generates synthetic data.
    %Using extensive experiments, we show that \texttt{Grasynda} improves forecasting accuracy and outperforms state-of-the-art time series data augmentation approaches.
    
    %Future work will pursue several promising directions. These include methodological improvements to \texttt{Grasynda}. Also exploring implementations with different graph construction or discretization techniques (e.g., visibility graphs)~\cite{Silva_2021} and utilizing the generation process methodology to stress-test forecasting networks by introducing intentional variability. 

    \bibliographystyle{splncs04}
    \bibliography{mybibliography}

    \end{document}